\documentclass{article}
\usepackage{spconf,amsmath,graphicx}
\usepackage{hyperref}
\usepackage{algorithmic}
\usepackage{subfigure}
\usepackage[font=small,skip=3pt]{caption}
\usepackage[linesnumbered,ruled,vlined]{algorithm2e}
\SetKwInput{KwInput}{Input}                
\SetKwInput{KwOutput}{Output}              

\title{A CNN regression model to estimate buildings height maps using Sentinel-1 SAR and Sentinel-2 MSI time series}

\name{Ritu Yadav \textsuperscript{1} \thanks{The research is part of the project ‘EO-AI4GlobalChange’ funded by Digital Futures.}, Andrea Nascetti \textsuperscript{1,2}, Yifang Ban \textsuperscript{1}}
\address{\textsuperscript{1}KTH Royal Institute of Technology (Sweden), \textsuperscript{2}University of Liège (Belgium)}
\begin{document}
%
\maketitle
\begin{abstract}
Accurate estimation of building heights is essential for urban planning, infrastructure management, and environmental analysis. In this study, we propose a supervised Multimodal Building Height Regression Network (MBHR-Net) for estimating building heights at 10m spatial resolution using Sentinel-1 (S1) and Sentinel-2 (S2) satellite time series. S1 provides Synthetic Aperture Radar (SAR) data that offers valuable information on building structures, while S2 provides multispectral data that is sensitive to different land cover types, vegetation phenology, and building shadows. Our MBHR-Net aims to extract meaningful features from the S1 and S2 images to learn complex spatio-temporal relationships between image patterns and building heights. The model is trained and tested in 10 cities in the Netherlands. Root Mean Squared Error (RMSE), Intersection over Union (IOU), and R-squared ($R2$) score metrics are used to evaluate the performance of the model. The preliminary results (3.73m RMSE, 0.95 IoU, 0.61 $R^2$) demonstrate the effectiveness of our deep learning model in accurately estimating building heights, showcasing its potential for urban planning, environmental impact analysis, and other related applications. 

\end{abstract}
\section{Introduction}
\label{sec:intro}
More than half of the world's population currently lives in cities. By 2050, an estimated 7 out of 10 people will likely live in urban areas. While cities contribute more than 80\% of global GDP they are also accountable for major energy consumption and carbon emission ~\cite{SDG_2022}. Urbanization monitoring is essential to assess its impact on the environment and support sustainable development. Accurate building height estimation plays an important role in urban planning, as it is an indicator of urban heat islands effect, population, energy consumption, and urban climate. 

Earth Observation (EO) has been highlighted as an effective tool for mapping large-scale human settlements. Several methodologies have been developed to extract building footprints in the last decades. Various large-scale global urban footprint data sets are available and widely used by the scientific community~\cite{marconcini2021understanding, hafner2022unsupervised}. However, these data sets are intrinsically two-dimensional and do not provide information on building height.
In recent years, some studies have tried to fill this gap and estimate building heights from satellite imagery. For example, \cite{li2020continental}, proposed a Random Forest (RF) regressor for continental-scale height mapping at 1 km spatial resolution. The authors used Landsat-8 OLI, Sentinel-1 SAR and various handcrafted spatial features along with auxiliary data. Reference data was derived from a combination of open street maps, government websites and commercial maps.
The authors of \cite{li2020developing} developed a VVH building height indicator from Sentinel-1 SAR data and estimated building heights at 500m resolution. The indicator was evaluated in major cities in the US with ICESat data as reference and achieved an RMSE of 1.5m.
\cite{esch2022world}, extended the World Settlement Footprint (WFS)~\cite{marconcini2021understanding}, including the building heights derived by the DSM collected by the TanDEM-X mission, and generated the WFS 3D data set at 90m resolution. The estimated building heights have been validated showing a promising accuracy with 6.01m RMSE score. However, it relies on a commercial DSM that is not easy to update frequently.
\cite{frantz2021national} presented a Support Vector Machine (SVM) regression model to derive building heights at 10m resolution with RMSE of 3.2m to 4.2m; the authors used a set of handcrafted spatial and temporal features from Sentinel-1 and Sentinel-2 time series as input to the model. The approach is tested in several cities in Germany using available ALS (Airborne Laser Scanner) data as a reference.
\cite{wu2023first} estimated building height for China at 10m resolution and achieved 6.1m RMSE. The authors used a combined approach from \cite{li2020continental} and \cite{frantz2021national} with additional ALOS PALSAR, WFS footprints and DEM data. The reference data is derived from Baidu map services with an assumption of each floor height to be 3m.

\begin{figure*}  
\centering  
\begin{subfigure}
  \centering
  \includegraphics[width=167mm, height=60mm]{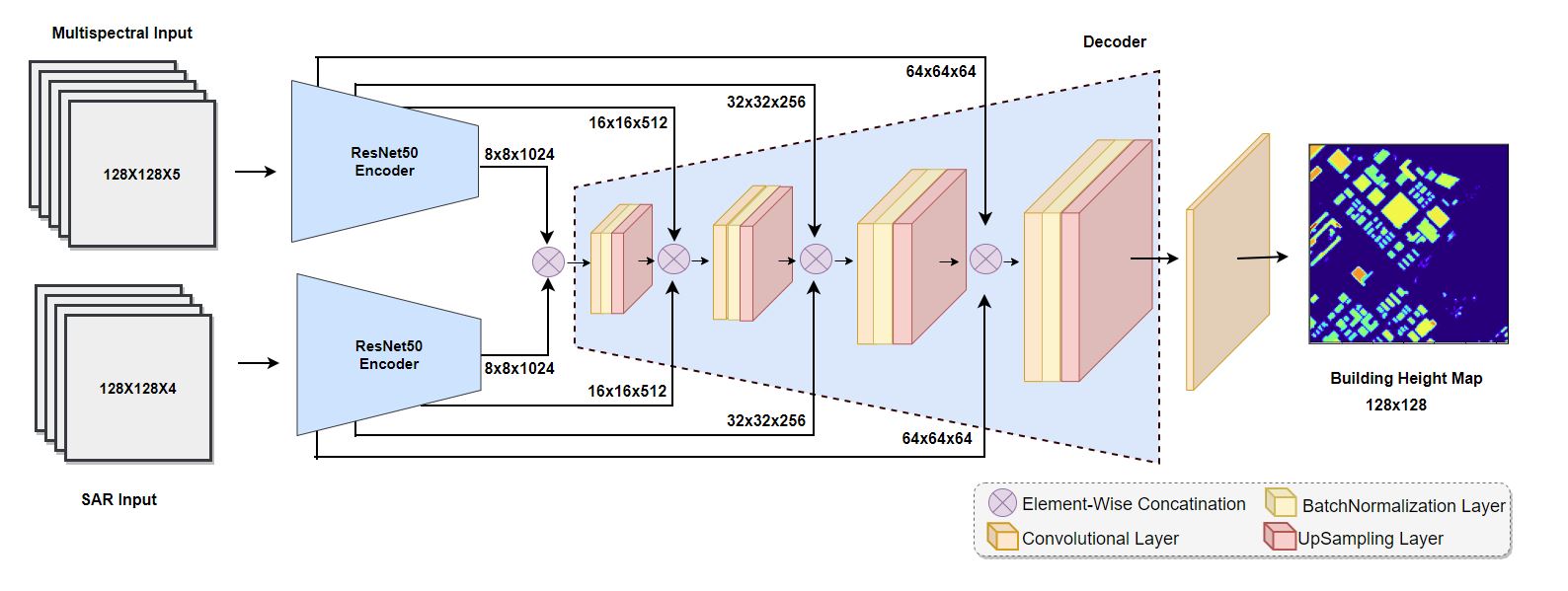}
\end{subfigure}
\vspace{-5px}
\caption{The proposed MBHR-Net for building height estimation.}
\label{network}
\vspace{-10px}
\end{figure*}
The aim of this study is to investigate a supervised Convolutional Neural Network (CNN) based regression model for estimating building height using only freely available Sentinel-1 SAR and Sentinel-2 MSI time series data. We frame the task of generating building height maps as a pixel-wise regression task, assuming the following: (1) zero-pixel values represent no buildings (as usual in urban footprint data), and (2) pixel values greater than 1 directly correspond to estimated building height. We developed a CNN regression model based on the U-Net architecture that takes Sentinel-1 and Sentinel-2 multi-temporal data and estimates building height at a 10m spatial resolution.

\section{Data Description}
\label{sec:Dataset}
The data used in this study include the Sentinel-1 Ground Range Detected (GRD) and Sentinel-2 MSI Level-2A. We collected data on ten largest cities in The Netherlands namely Amsterdam, Rotterdam, The Hague, Utrecht, Eindhoven, Groningen, Breda, Tilburg, Nijmegen and Almere. 
For reference, we used 3D Bag data developed by the 3D geoinformation Group of the Technical university of Delft. The database is completely open source and automatically generated from the Buildings and Addresses Register (BAG) providing 2D footprints and from the National Altimetric Model (AHN) derived from ALS data. The database contains multiple Levels Of Detail (LOD) building models (LOD1.2, LOD1.3 and LOD 2.2). We selected LOD1.3 data for this study.
We created non-overlapping tiles of size 128x128 pixels within the administrative boundaries of the 10 metropolitan areas provided by the European Environment Agency. Corresponding to each tile we collected S1 and S2 time series using Google Earth Engine's python API \cite{GORELICK201718} and collected LOD1.3 data from 3D BAG database. For S2 MSI data, we generated monthly cloud-free composites and downloaded 5 bands (Red, Green, Blue, NIR and SWIR). For S1 SAR data we first computed the monthly average to reduce the speckle for both ascending and descending orbits, then downloaded 4 bands (VV, VH polarizations for both orbits). The S2 data contains 12 images per tile (one image for each month) and the year is matched with the acquisition of AHN data. The input data is downloaded at 10m spatial resolution, and the reference data are rasterized and resampled to match 10m. We divided the dataset into training and test set using an 80-20 ratio, resulting in 1,737 training samples and 434 test samples.

\section{Methodology}
\subsection{MBHR-Net Architecture}
\label{sec:Network}
Figure \ref{network} shows the architecture of the proposed MBHR-Net, consisting one branch for learning multispectral features of S2 images and another branch for learning SAR backscatter features of S1 images. The S2 branch takes a five-channel input (red, green, blue, NIR and SWIR bands) and the S1 branch takes a four-channel input (VV, VH for both descending and ascending orbits).

We adopted U-Net architecture, a widely used encoder-decoder based segmentation network. An encoder compresses the salient features (feature maps) of the input images and a decoder upsamples the compressed features to predict output of same size as input. The encoder block has a number of repetitions (level) of the sequence; convolutional layer, maxpooling layer and batch normalization layer. The decoder has a sequence of convolutional layers, upsampling layer and back normalization layer to output a segmentation map.

Our MBHR-Net contains two encoders one in each branch to learn different modality features separately. We adopted ResNet50 with four levels as encoder. The output feature maps of each level is of different size capturing different semantics. From each encoder, four feature maps are extracted. The size of the feature maps are (64x64), (32x32), (16x16) and (8x8). These multiscale features are fused using element-wise concatenation operation. Via skip connection, the fused features maps are combined with the same size decoder layers. These skip connections help the decoder network to condition not only on the latent representation but also on intermediate representations of the encoder, which lead to fine-grained details in predictions\cite{yadav2022unsupervised, ma2019deep}. The combined feature maps are upsampled and processed through decoder layers. At the end of the decoder network we use a convolutional later with (1x1) kernel and $ReLU$ activation function making it a regression layer.

\subsection{Augmentation strategy and Training}
For one reference patch we have 24 input images i.e. 12 time series images for each modality. These 12 images capture surrounding features in different months. We assume that in a 12-month period, there are negligible changes in building heights but the season changes surrounding conditions. Therefore, the 12-month images can be treated as augmented images, creating 12 augmented pairs. With augmentation, the size of training samples increased from 1734 to 1734x12.
As training losses we used a weighted combination of two regression losses, Mean Squared Error loss (MSE) and Cosine Similarity (CS) loss given in equation Eq. \ref{eq:mse} -- \ref{eq:cs}. We used 0.8 weight for CS loss and 0.2 for MSE loss.
\begin{equation} 
    \label{eq:mse}
    \begin{aligned}
    MSE = (y_{true} - y_{pred})^{2}    
    \end{aligned}
\end{equation}
\begin{equation} 
    \label{eq:cs}
    \begin{aligned}
    CS = -\sum_{}^{}\left\| y_{true} \right\|2 * \left\| y_{pred} \right\|2
    \end{aligned}
\end{equation}

We trained the model for 100 epochs with batch size 4, adam optimizer and 0.0001 as initial learning rate. For better convergence, the learning rate is decayed until 0.00001. The decay steps are controlled with the "reduce on plateau" method. The code is implemented in Keras and the model is trained for 6 hours on Google colab GPU.

\section{Results and Evaluation}
\label{sec:eval}
The predicted building height maps MBHR-Net are first filtered using building footprints obtained by binarizing the reference height maps. For binarization, a pixel is set to be building pixel (1.0) if the pixel value is $>$ 1.0 otherwise set to no building (0.0 value). The filtered building height maps are evaluated using two metrics Root Mean Square Error (RMSE) and $R^{2}$ score given in Eq. \ref{eq:rmse}, \ref{eq:rscore}, where $n$ is the number of validation samples, $BH_{est, i}$ is the estimated value of the height of the building and $BH_{ref, i}$ is reference building height. RMSE indicates the accuracy of predicted heights with respect to reference and $R^{2}$ score estimates the model effectiveness in learning variance in the building heights.
\begin{equation} 
    \label{eq:rmse}
    \begin{aligned}
    RMSE = \sqrt{\frac{\sum_{i=1}^{n}(BH_{ref, i} - BH_{pred, i})^{2}}{n}}
    \end{aligned}
\end{equation}
\begin{equation} 
    \label{eq:rscore}
    R^{2} = 1- \frac{(n-1)\sum_{i=1}^{n}(BH_{ref, i} - BH_{pred, i})^{2}}{(n-2)\sum_{i=1}^{n}(BH_{ref, i} - BH_{pred, i})^{2}}
\end{equation}
For accurate building height estimation, it is important to ensure the alignment of predicted buildings with the reference. We evaluated building alignment using well known metric Intersection over Union (IoU). RMSE and $R^{2}$ scores are calculated using reference and network predictions directly whereas IoU is calculated on binarized (building and no building) references and predictions.

A high-quality building height estimation model is characterized by a low RMSE, a high IoU, and a high $R^{2}$ score. These metrics provide insight into different aspects of the model's performance. The RMSE, IoU and $R^{2}$ scores of the proposed MBHR-Net are 3.73 m, 0.95 and 0.61. The RMSE score of 3.73 m suggests that, on average, the model's accuracy is approximately one floor showing a remarkable accuracy considering the spatial resolution of S1 and S2 imagery (10-20 m). The IoU score of 0.95 indicates that the predicted height regions overlap significantly with the ground truth regions, demonstrating the model's ability to precisely identify buildings. The $R^{2}$ score is 0.61, suggesting that the model can explain a substantial portion of the variance in building heights but the model may have limitations in accurately capturing certain factors or complexities affecting building heights.
\begin{table}[htbp]
\caption{Performance metrics on test set.}
\begin{center}
\resizebox{.9\columnwidth}{!}{%
\begin{tabular}{c|c|c|c}
 & RMSE (m) & $R^{2}$  & IoU \\ \hline \hline 
mean $\pm $ std & 3.73$\pm $2.01 & 0.61$\pm $0.12 & 0.95$\pm $0.10    \\ 
\end{tabular}%
}
\vspace{-15px}
\end{center}
\label{SOTA_comp2}
\end{table}

Figure \ref{network} shows a scatter plot between predicted heights and reference heights of all pixels with values greater than 1 in the test set. In the scatterplot we reported that the model shows a fair correlation between reference heights and predictions. But there is an overall underestimation of heights, which we aim to improve in our future works.
\vspace{-3px}
\begin{figure}[htbp]
\caption{Scatter plot of the predicted and reference height values. The evaluation is on test set and the height values are in meters (m).}
\centerline{\includegraphics[height=70mm]{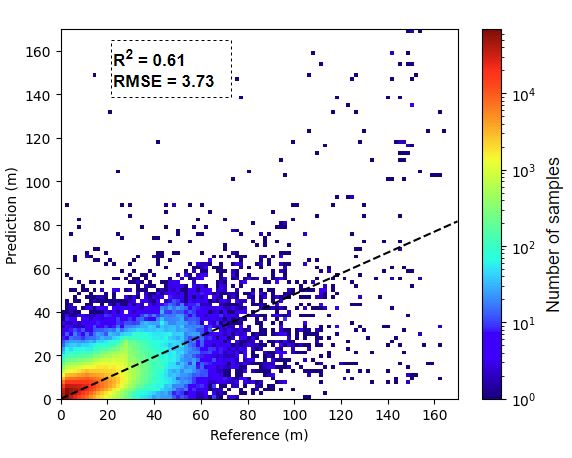}}
\label{scatter_plot}
\end{figure}
\vspace{-10px}

\begin{figure*}  
\centering  
\begin{subfigure}
  \centering  
  \includegraphics[width=150mm, height=100mm]{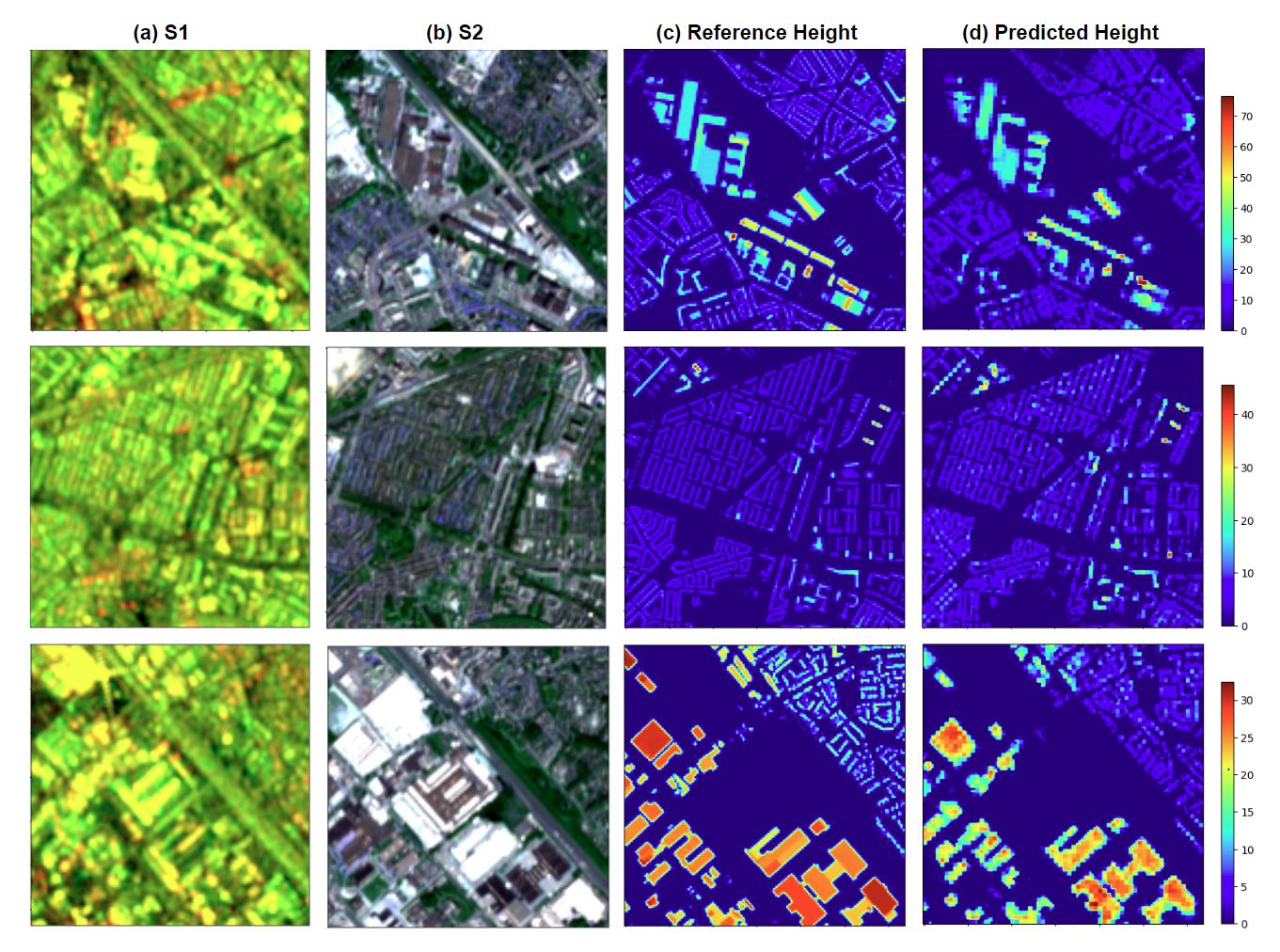}
\end{subfigure}
\caption{Result Samples. Three sample results are visualized in 3 rows. From left to right, Sentinel-1 SAR, Sentinel-2 MSI, Reference Building Height and Predicted Building Height (in meters) from our MBHR-Net.}
\vspace{-10px}
\label{data_samples}
\end{figure*}

We also present a few test samples for qualitative evaluation (see Figure \ref{data_samples}). The first two rows show good height estimation examples where the predicted heights are accurate with few underestimations. The third sample shows comparatively less accurate height estimations with error ranges of 0 to 4 meters. This is possibly an example of the type of complexity which is not accurately learned by the model. The high IOU score of the model can be verified from all samples showing a good alignment between the predicted buildings and the ground truth buildings, including the boundary areas.

\section{Conclusion}
\label{sec:conclusion}
In this study, we developed a deep learning model for building height estimation using combined Sentinel-1 SAR and Sentinel-2 MSI time series. The performance evaluation of MBHR-Net demonstrated promising accuracy in both height estimation, with an RMSE of 3.73m, and building footprint delineation, with a 95\% IoU. These results indicate the potential of MBHR-Net for estimating building heights with accurate building footprint delineation. The potential future investigation directions are to expand the data set to include different geographic regions, building types, incorporate additional data sources, and develop a more advanced deep learning model to handle the complexity of such large data to build a more generalized approach.
\bibliographystyle{IEEEbib}
\bibliography{refer}

\end{document}